\documentclass[sn-apa]{sn-jnl}

\usepackage[utf8]{inputenc}
\usepackage[T1]{fontenc}
\usepackage[english]{babel}

\usepackage{lmodern}
\usepackage{anyfontsize}

\usepackage{natbib}
\bibpunct{(}{)}{;}{a}{,}{;}

\usepackage{graphicx}%
\usepackage{multirow}%
\usepackage{amsmath,amssymb,amsfonts}%
\usepackage{amsthm}%
\usepackage{mathrsfs}%
\usepackage[title]{appendix}%
\usepackage{textcomp}%
\usepackage{manyfoot}%
\usepackage{booktabs}%
\usepackage{algpseudocode}%
\usepackage{listings}%
\usepackage{rotating}%

\usepackage[pdftex,dvipsnames,svgnames,hyperref,fixpdftex,table]{xcolor}

\usepackage{tabularx, lscape, longtable, booktabs, multicol, multirow, dcolumn}
\usepackage{makecell}

\usepackage{amsmath, amsfonts, amssymb, amsthm, mathrsfs, mathtools}

\usepackage[pagewise]{lineno}

\usepackage{xspace}

\usepackage[bottom]{footmisc}

\usepackage[algoruled,vlined,linesnumbered,norelsize]{algorithm2e}

\usepackage[normalem]{ulem}

\usepackage{adjustbox}
\usepackage{comment}
\usepackage{chngpage}

\newcommand{\shakeiterv}{\text{\slshape I}_s}        
\newcommand{\resetmul}{\text{\slshape R}_m}    
\newcommand{\calP}{\ensuremath{{\mathcal P}}}

\begin{document}




\title[Article Title]{Early years of Biased Random-Key Genetic Algorithms: A systematic review}

\author[1]{\fnm{Mariana~A.} \sur{Londe}}\email{mlonde@aluno.puc-rio.br}

\author[1]{\fnm{Luciana~S.} \sur{Pessoa}}\email{lucianapessoa@puc-rio.br}

\author[2]{\fnm{Carlos~E.} \sur{Andrade}}\email{cea@research.att.com}

\author*[3]{\fnm{Mauricio~G.~C.} \sur{Resende}}\email{mgcr@uw.edu}

\affil[1]{\orgdiv{Department of Industrial Engineering}, \orgname{PUC-Rio}, \orgaddress{\street{Rua Marqu\^es de S\~ao Vicente, 225, G\'avea}, \city{Rio de Janeiro}, \postcode{22453-900}, \state{RJ}, \country{Brazil}}}

\affil[2]{\orgdiv{AT\&T Labs Research}, \orgname{AT\&T}, \orgaddress{\street{200 South Laurel Avenue}, \city{Middletown}, \postcode{07748}, \state{NJ}, \country{USA}}}

\affil*[3]{\orgdiv{Industrial \& Systems Engineering}, \orgname{University of Washington}, \orgaddress{\street{3900 E Stevens Way NE}, \city{Seattle}, \postcode{98195}, \state{WA}, \country{USA}}}

\abstract{
This paper presents a systematic literature review and bibliometric analysis focusing on Biased Random-Key Genetic Algorithms (BRKGA). BRKGA is a metaheuristic framework that uses random-key-based chromosomes with biased, uniform, and elitist mating strategies alongside a genetic algorithm. This review encompasses around~250 papers, covering a diverse array of applications ranging from classical combinatorial optimization problems to real-world industrial scenarios, and even non-traditional applications like hyperparameter tuning in machine learning and scenario generation for two-stage problems. In summary, this study offers a comprehensive examination of the BRKGA metaheuristic and its various applications, shedding light on key areas for future research.
}%
%

\maketitle



\section{Introduction}
\label{Section:Introduction}

Pioneered by \citet{Holland1975:genetic_algorithm}, genetic algorithms have demonstrated remarkable efficacy in tackling complex optimization problems. These algorithms are population-based metaheuristics inspired by the principles of natural selection and survival of the fittest, which turns them into powerful tools for exploring expansive solution spaces. Their versatility and efficiency have led to their widespread application across various domains, addressing problems ranging from discrete and combinatorial to nonlinear and derivative-free optimization problems.

A popular variant of genetic algorithms, the Biased Random-Key Genetic Algorithms (BRKGA), have been applied to several optimization problems, such as scheduling \citep{Araujo2015:parallel_assembly_lines,%
Yu2023:supply_scheduling,%
Andrade2019:scheduling_software_cars,%
Maecker2023:unrelated_parallel_machine_scheduling, Cunha2018:rescue_allocation},
complex network design 
\citep{Reis2011:OSPF_DEFT_routing_network_congestion,%
Ruiz2015:capacitated_minimum_spanning_tree,%
Raposo2020:network_reconfiguration_energy_loss,%
Andrade2022:PCI_Assignment_practical_opt}, 
facility location 
\citep{Johnson2020:cover_by_pairs,%
Mauri2021:multiproduct_facility_location,%
Villicana2022:mobile_labs_covid19_testing}, cutting and 
packing \citep{Goncalves2011:multi_pop_constrained_2d_orthogonal_packing,Goncalves2020:two_dimensional_cutting_defects},
clustering \citep{Andrade2014:Evol_Alg_Overlapping_Correlation_Clustering,Fadel2021:statistical_disclosure_control}, 
vehicle routing \citep{Andrade2013:Evolutionary_Algorithm_kIMDMTSP,%
Lopes2016:hub_location_routing}, graph-based problems \citep{Londe2022:root_sequence_index,Lima2022:matheuristic_broadcast_time}, and machine learning \citep{%
Caserta2016:data_fine_tuning,%
Paliwal2020:Reinfored_BRKGA_Comp_Graphs}, among others. This metaheuristic was formally defined in \citet{Goncalves2011:BRKGA}, although early elements were first introduced in \citet{%
Beirao1997:FirstBRKGA,%
Buriol2005:weight_setting_problem_OSPF_routing,%
Ericsson2002:Genetic_alg_OSPF,%
Goncalves1999:BRKGAFirst,%
Goncalves2002:hybrid_assembly_line_balancing%
}.

The primary distinguishing characteristic of BRKGA is its problem-agnostic approach. Unlike many meta\-heuristic algorithms where the optimization mechanism is closely tied to the problem's structure, BRKGA operates within a general framework and standard representation of solutions, which minimizes the need for continual redevelopment or coding of framework details. In this solution approach, the population resides within a half-open unit hypercube of dimension~$n$, and each solution or individual is denoted by a point in~$(0,1]^n$, termed as a \emph{chromosome}. This representation, introduced in \citet{Bean1994:random_keys}, frees the method from dependence on the specific problem it addresses, allowing for code reuse.
Such a strategy resembles 
modern machine learning algorithms: the knowledge representation is built over
a normalized matrix, and the so-called kernel functions are applied to measure
distances between points in that normalized
space \citep{Hofmann2008:Kernel_methods}. In a BRKGA, instead of having a kernel,
we have a \emph{decoder function} $f:(0,1]^n \to \mathcal{S}$ that maps
individuals from the BRKGA space to the problem solution space~$\mathcal{S}$.
Indeed, the decoder not only builds an actual solution from a chromosome but
also computes the solution value(s) used by the BRKGA as a measure of the quality
or fitness of the individual. We may see the decoder as the function that
computes the ``norm'' of a solution in the $(0,1]^n$ space. Such representation
allows a BRKGA to keep all evolutionary operators within the $(0,1]^n$ space, and
therefore, custom operators based on the problem structure are unnecessary.
This allows for fast prototyping and testing, thus reducing development costs.

Another notable characteristic of a BRKGA is its rapid convergence to high-quality solutions. This achievement stems from the inclusion of the double elitism mechanism within the evolutionary process of BRKGA. First, BRKGA 
hands over a subset of \emph{elite} individuals from one generation to the next, according to some performance metric (in general, the value(s) of
the objective function of the problem). The elite individuals are the best solutions for the current
generation. Excluding the very first iterations, a BRKGA will always have a set of
high-quality solutions in its population. This
behavior contrasts with traditional genetic algorithms, which generally rebuild
the whole population every generation. Second, in the basic BRKGA, the mating
process occurs between a uniformly chosen individual from the elite set and a
uniformly chosen individual from the remaining population. The combination of
such individuals is biased towards the elite individual, using uniform
crossover with probability~$\rho > 0.5$. Thus, there is a greater chance
of retaining substructures of a good solution while still allowing the insertion of substructures
of a not-so-good solution. 

The downside of double elitism is a swift convergence to local optima. BRKGA addresses these under-performing issues by incorporating strategies to increase population diversity, such as the introduction of mutants (random solutions), shaking \citep{Andrade2019:flowshop_scheduling}, population reset \citep{Toso2015:C++_app_BRKGA}, and others. For a detailed explanation of the evolutionary mechanism and variants of BRKGA, see \citet{Londe2024:review_brkga}. Notwithstanding, BRKGA can provide high-quality solutions within short computational times, rendering it suitable for various industrial applications.

Due to these attributes, the use of BRKGA has grown significantly in recent years. In this survey, we present a systematic literature review
(SLR) and bibliometric analysis of the existing literature regarding BRKGA to
discern the most studied problems, principal modifications to the framework,
and most influential authors and works since its conception. An SLR aims to
synthesize the accumulated body of knowledge about the relations of interest
and to determine where gaps exist \citep{Gligor2012:logistics_SLR}. Meanwhile,
a bibliometric analysis is the quantitative study of bibliographic material
offering an overview of a research field classified by papers, authors, and
journals \citep{Merigo2017:bibliometric_analysis}. 
Approximately~250 academic articles were reviewed for this survey.
We expect that readers will gain a comprehensive and expansive understanding of BRKGA applications, aiding them in their future pursuits.


The remainder of this article is structured as follows. In Section~\ref{Section:Features} we present the main features and characteristics of BRKGA.
Section~\ref{Section:Methodology} describes the methods used in the literature
review and bibliometric analysis. Section~\ref{Section:Results:Citation}
presents the results from citation analysis. Section~\ref{Section:Results:Cocitation} presents the backdrop of BRKGA research obtained by the co-citation analysis. In Section~\ref{Section:Results:Coword}, we comment on the thematic evolution of the corpus of literature with a co-word analysis. Finally, in Section~\ref{Section:Conclusion}, we make concluding
remarks.

\section{Main features of BRKGA}
\label{Section:Features}


As \citet{Goncalves2011:BRKGA} describes, Biased Random-Key Genetic Algorithms (BRKGA) belong to the category of Genetic Algorithms (GA), which are a class of combinatorial optimization algorithms that are inspired by Darwin's theory of evolution. \citet{Holland1975:genetic_algorithm} introduces this category, which considers a possible solution to a combinatorial optimization problem as an individual inside a population of potential solutions. Each individual is represented by a chromosome, itself characterized as a string or vector of genes. 

In BRKGA, a gene is a real number belonging to the interval~$(0,1]$. \citet{Bean1994:random_keys} introduces this concept, which frees the algorithm from a problem-specific representation and allows for a generalization of the framework. This characteristic is prevalent in several features of the BRKGA, from traditional operators such as crossover to novel components such as implicit path relinking \citep{Andrade2021:BRKGA_MP_IPR}.

The classic BRKGA framework considers three evolutionary procedures whose application obtains new $p$ population(s) of $|\calP|$ solutions. The \emph{reproduction} operator copies the best potential solutions to the next generation. This mirrors Darwin's idea of the survival of the strongest, fittest individuals, who are capable of transmitting their characteristics to the next generations. The set of best solutions is known as the \emph{elite set} and corresponds to a percentage $\calP_e\%$ of the complete population.

Likewise, the weakest individuals do not transmit their features to the new population, being instead replaced by novel individuals obtained from the \emph{mating} and \emph{mutant generation} processes. The \emph{mutant generation} operator creates new random individuals and inserts those in the new population(s). Those are obtained in the same way as the individuals of the initial population, with genes as random--keys obtained from real values on a uniform distribution belonging to interval~$(0,1]$. The amount of mutants is determined by a percentage $\calP_m\%$ of the complete population.

Meanwhile, the \emph{mating} procedure is the combination of solutions from the elite and non-elite sets. In classic BRKGA, it is a parameterized uniform crossover \citep{Spears1991:multi_point_crossover}. A biased coin toss is performed for each gene of the offspring, to determine which parent will contribute. There is a $\rho > 0.5$ probability of selecting the fitter parent, which contributes to the propagation of characteristics of the most promising solutions throughout the evolutionary process. This process introduces $|\calP| - |\calP^e| - |\calP^m|$ individuals into the next population.

In BRKGA, the fitness of a given solution is obtained by the decoding process. This operator is a part of the framework that must be customized for each problem. A decoder translates the vector of random--keys to a feasible solution and returns its fitness value. 



Since its formal introduction in \citep{Goncalves2011:BRKGA}, several features were added to the BRKGA framework. We considered a feature as a strategy adapted specifically to the algorithm and applied since its inception. 

One such feature is the use of parallel $p > 1$ populations, known as the island model \citep{Whitley1999:island_model}. It is based on another characteristic of Darwin's evolution theory, in which isolated populations have divergent evolutionary paths. This feature also comprises the migration of individuals after a $g$ generations. This procedure improves individual variability \citep{Andrade2021:BRKGA_MP_IPR}, thus preventing premature convergence \citep{Pandey2014:premature_convergence}, and has been frequently added to the BRKGA framework since the introduction by \citet{Toso2015:C++_app_BRKGA} of the C++ API.

Another common addition also introduced in this API is the reset operator \citep{Toso2015:C++_app_BRKGA,Whitley1999:island_model}. A full population restart may be needed if the algorithm cannot escape local optima for a high number of generations. Thus, this operator re-initializes all individuals in all populations with randomly generated genes, which destroys the benefits of convergence but may explore new locations in the solution space. A variant of this is the shaking procedure \citep{Andrade2019:flowshop_scheduling}, which is a partial restart of the populations. The shaking operator applies random perturbations to the genes of the elite individuals and re-initializes the non-elite chromosomes. Thus, it preserves some of the benefits of convergence while increasing population diversity and potentially escaping local optima. Shaking is usually performed after $\shakeiterv$ iterations and resetting occurs after $\resetmul \cdot \shakeiterv$ generations without improvement of the best solutions.

A less common feature is the online parameter tuning, first introduced to the BRKGA framework by \citet{Chaves2018:capacitated_centered_clustering_local_search}. This approach updates the parameters relative to crossover probability $\rho$, population size $|\calP|$, elite $\calP_e\%$ and mutant $\calP_m\%$ percentages, and the maximum number of generations at each iteration. This feature carefully balances exploration and exploration strategies alongside the evolutionary process. Later, a Q-Leaning reinforced learning algorithm was used to tune the same parameters \citep{Schenekemberg2022:dial_a_ride}, with an exponential decrease of diversification during the evolution.

More recently, \citet{Andrade2021:BRKGA_MP_IPR} introduced two new features to the BRKGA framework. The first is the multi-parent crossover, in which $\pi_t$ parents are selected for the mating process, with $\pi_e$ coming from the elite set and the remainder of the non-elite set. To continue with the elitist strategy of the original BRKGA, the genes are selected with a fitness ranking-based bias function $\Phi(r)$\citep{Bresina1996:stochastic_sampling} with an increased chance of the fitter parents contributing to the offspring. 

The second feature is the implicit path-relinking procedure (IPR). Classic path-relinking explores the neighborhood obtained in the path between two distinct, distant solutions. Due to the distance element, the use of this method is usually problem-dependent \citep{Glover1997:PR,Ribeiro2012:PR}. The implicit variant is constructed inside the BRKGA random-key framework alongside parallel populations so that the base and guide solutions have distances of at least $md$, as calculated by either Kendall tau or Hamming distances. The method is defined by the type $typ$ of IPR selected, which depends on the interpretation of the chromosomes. This procedure may select the best solutions of each population or randomly choose one of the elite individuals as the base or guide.

\section{Methodology}
\label{Section:Methodology}



\subsection{Systematic review and basic statistics}
\label{Section:Methodology:BasicStatistics}

\citet{Thome2016:SLR_OM} details eight steps to make a systematic literature
review in the area of operations management: (i) Planning
and formulating the problem; (ii) Searching the literature; (iii) Data
gathering; (iv) Quality evaluation; (v) Data analysis and synthesis; (vi)
Interpretation; (vii) Presenting the results; and (viii) Updating the review.

The first step, planning and formulating the problem, is composed of the definition of the review's scope, the detailing and definition of the research topic, and the outlining  of the subsequent research questions (RQ):
\begin{description}
\item[RQ1:] Who are the most influential researchers for this algorithm?
\item[RQ2:] Which are the most influential papers for the BRKGA framework?
\item[RQ3:] Which are the main themes present in BRKGA studies and how did they evolve?
\end{description}

We adhere to the seven-step method detailed in \citet{Thome2016:SLR_OM} to search the literature for the second step. Those steps are (i) the selection of the database, (ii) the definition of keywords, (iii) the review of abstracts, (iv) the definition of criteria for inclusion and exclusion of works, (v) full-text review, and (vi) backward and (vii) forward search in the selected works.

We chose the Scopus database \citep{Baas2020:scopus,Scopus2022} for this review, which is one of the largest
curated abstract and citation databases, with over~45,000 journal titles from~7,000 publishers worldwide. 
\citet{Singh2021:WoS_Scopus} observes that it is less restrictive in the
selection of titles than its competitor Web of Science and has a bias toward the fields of technology and engineering.

When defining the keywords, one must ensure their broadness and specification not to restrict the number of studies and to bring only works related to the topic \citep{Cooper2015:meta_analysis}. The keywords used were ``BRKGA'' or ``Biased random key genetic algorithm,'' resulting in~267 works. The limitation to peer-reviewed articles and conference papers in the English language resulted in~231 works. We considered as inclusion criteria that the work must detail and use a Biased Random-Key Genetic Algorithm to solve a practical or theoretical problem, not only cite or compare another approach's results with those from a BRKGA. We also decided to include articles with a genetic algorithm with random keys and biased mating to not exclude older titles from before the formal definition of BRKGA in~2011. The full-text review of the initial works excluded two studies. Then, we employed backward and forward snowball searches. The former is a review of the literature cited in the articles yielded from the initial search. At the same time, the latter is a review of additional articles that cite those retrieved \citep{Webster2002:lit_review}. Due to the difficulty in obtaining older BRKGA works from before the introduction of the term with the previous keywords, we added the keywords ``hybrid genetic algorithm'' and ``random-key genetic algorithm'' to the backward search, which are terms used to refer to the algorithm before~2011. After applying the keywords to the new registers alongside the exclusion criteria and eliminating all works with non-related abstracts, we end up with~253 articles published before December~31st,~2023.


\begin{figure}
    \centering
    \includegraphics[scale = 0.7]{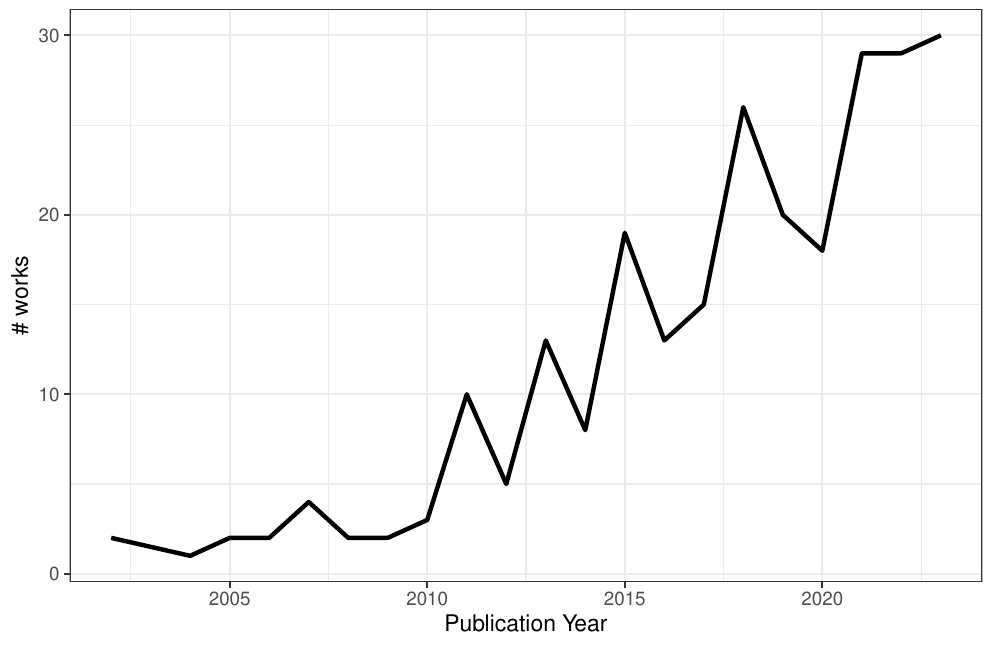}
    \caption{Number of BRKGA articles published per year.}
    \label{Figure:articles_per_year}
\end{figure}

Figure~\ref{Figure:articles_per_year} shows the number of published articles
per year for those~253 papers. One can note a severe increase of
published works after~2011, i.e. after \citet{Goncalves2011:BRKGA} formally
defined the metaheuristic. In fact,~89\% of the~253 articles were published in
the 2012--2023 period, with an average of~19 papers per year. Those papers include over 100 sources and 450 authors, with an average of three authors per article and 23 citations per document. 


Table~\ref{Table:journals_no_citations} presents the twenty sources with the highest amount of papers. One may note that those contribute with~83\% of all
citations inside the database -- and that a high amount of articles does not
necessarily correspond to a high contribution of citations. This may be seen with the source \textit{Lecture Notes in Computer Science}, with the highest amount of articles and yet only~2\% of citations. Generally, those sources with low citations-per-paper tend to be congress proceedings, such as the aforementioned \textit{Lecture Notes} and the three \textit{Proceedings of IEEE CEC}.

\begin{table}[htb]
    \centering
     \caption{20 sources with the highest number of published articles. The \% of citations is observed inside the database.}
    \begin{tabular}{lcc}
        \hline
        \textbf{Source} & \textbf{No. p.} & \textbf{\% cit.}\\
        \hline
        Lecture Notes in Computer Science & 23 & 2\\
        European Journal of Operational Research & 17 & 23\\
        Computers and Operations Research & 15 & 12\\
        International Transactions in Operational Research & 15 & 5\\
        Applied Soft Computing & 9 & 9\\
        Computers and Industrial Engineering & 9 & 3\\
        Journal of Heuristics & 8 & 13\\
        Optimization Letters & 6 & 2\\        
        Expert Systems with Applications & 5 & 2\\
        Journal of Combinatorial Optimization & 5 & 5\\
        Proceedings of 2018 IEEE CEC & 4 & 2\\
        Proceedings of 2021 IEEE CEC & 4 & 3\\
        Journal of Global Optimization & 4 & 1\\
        Networks & 4 & >1\\
        Proceedings of 2022 IEEE CEC & 3 & 1\\
        Proceedings of GECCO 2019 & 3 & >1\\
        International Journal of Production Research & 3 & >1\\
        Mathematics & 3 & >1\\
        Pesquisa Operacional & 3 & >1\\
        RAIRO - Operations Research  & 3 & >1\\
        \hline
    \end{tabular}   
    \label{Table:journals_no_citations}
\end{table}


The third step of the review can be fulfilled by using a computer template and programs to calculate amounts and frequencies in the database of papers. Meanwhile, for step four, quality evaluation, the restriction of only considering peer-reviewed articles and complete papers presented at international conferences is a criterion that helps ensure the quality of the resulting studies.

For the fifth step, data analysis and synthesis, we perform a qualitative content analysis with an inductive approach \citep{Seuring2012:content_analysis_LR} and quantitative co-citation and co-word analysis described in Subsection~\ref{Section:Methodology:Bibliometric}. The sixth step, interpretation, comprises qualitative research synthesis coupled with indicators from the bibliometric analysis. The seventh step is the presentation of the results obtained in the previous steps. Finally, the eighth step, updating the review, does not apply to this study.


\subsection{Bibliometric analysis}
\label{Section:Methodology:Bibliometric}

Several indicators can be used for bibliometric analysis. In this review, we use co-citation and co-word analyses. Co-citation analysis groups different papers that are cited by the same source. For example, if papers~A and~B cite paper~C, then paper~C is co-cited by both papers~A and~B~\citep{Boyack2010:cocitation_analysis}. A co-citation network indicates subject relatedness, as it is likely that papers citing the same papers focus on similar subjects \citep{Thome2016:supply_chain_projects}. Similarly, a co-word analysis uses keywords to study the conceptual structure of a research field \citep{Callon1991:coword_analysis_diagram,Cobo2011:mapping_tools_review}.

For those analyses, we used the R bibliometrix package \citep{Aria2017:bibliometrix} that is available as free-ware for non-profit academic use. It was used to prepare the matrix of co-citations, clusters, and network analysis, along with the dynamic maps of co-word analysis of themes using Callon's thematic bipartite diagrams \citep{Callon1991:coword_analysis_diagram}.

\begin{figure}
    \centering
    \includegraphics[scale = 0.6]{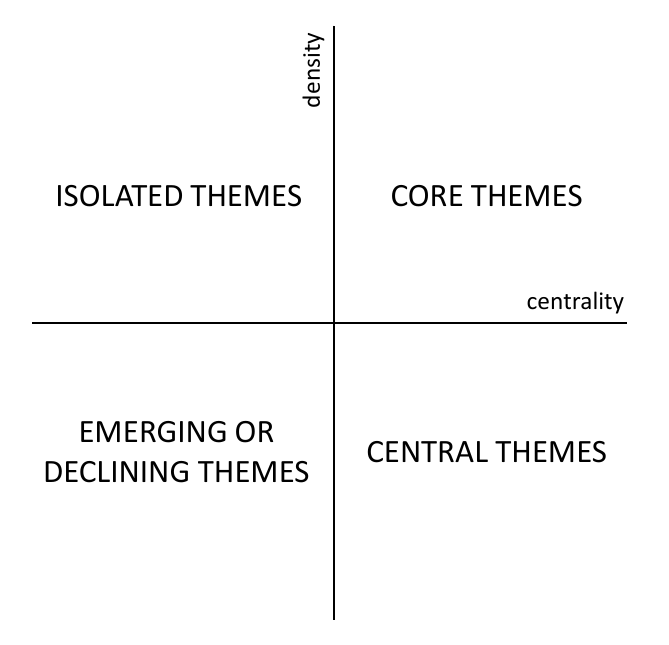}
    \caption{Callon's diagram. Adapted from \citet{Callon1991:coword_analysis_diagram}}
    \label{Figure:Callons_diagram}
\end{figure}

Callon's diagram can be seen in Figure~\ref{Figure:Callons_diagram}. In it, the
top right quadrant represents the core themes of the research area, with high
density and high centrality (which are defined in what follows). The central themes are also crucial for the area
but are not yet well researched, with opportunities for new studies. Emerging
or declining themes have low centrality and density, i.e., they are not
strongly related to other themes and are not well represented in the research area.
Finally, isolated themes are well-researched areas with high centrality
and, thus, of marginal value for the field. Usually, they represent classical
themes in the field.

The co-occurrence of keywords is measured by the similarity index $e_{ij} =
c_{ij}^2/c_{i}c_{j}$, in which~$c_{ij}$ is the number of works in which both
keywords $i$ and $j$ are featured, and~$c_{i}$ and~$c_{j}$ are, respectively, the numbers of works in which each keyword $i$ and $j$
occurs \citep{Cobo2011:mapping_tools_review}. The clusters are formed with the
``simple center
algorithm'' \citep{Cobo2011:mapping_tools_review,Thome2016:supply_chain_projects},
with centrality and density calculated as linear functions of the similarity
index. Centrality is calculated as $C = 10 \cdot \sum e_{kh}$, where~$k$ is a
keyword belonging to the theme and~$h$ a keyword belonging to the other themes,
as a measure of the interaction among the network of keywords. Meanwhile,
density is calculated as $D = 100 \cdot \sum e_{ij}/w$, where~$i$ and~$j$ are
keywords belonging to the theme and~$w$ is the total number of keywords in the
theme. It measures the internal strength of the network or theme.

\section{Citation analysis}
\label{Section:Results:Citation}

To answer RQ1, we present Table~\ref{Table:Results:Authors}, with the~10 most prolific authors. In this table, the results are indicated inside of the database of~253 articles. One may note that the author with the most papers (Resende, M.G.C.) has the highest total citations among the works in the database. We calculated a Pearson correlation coefficient of 0.87 between the number of papers and the total number of citations inside the database, showing a strong correlation between those two indicators.

\begin{figure}[htb]
    \centering
    \includegraphics[scale = 0.3]{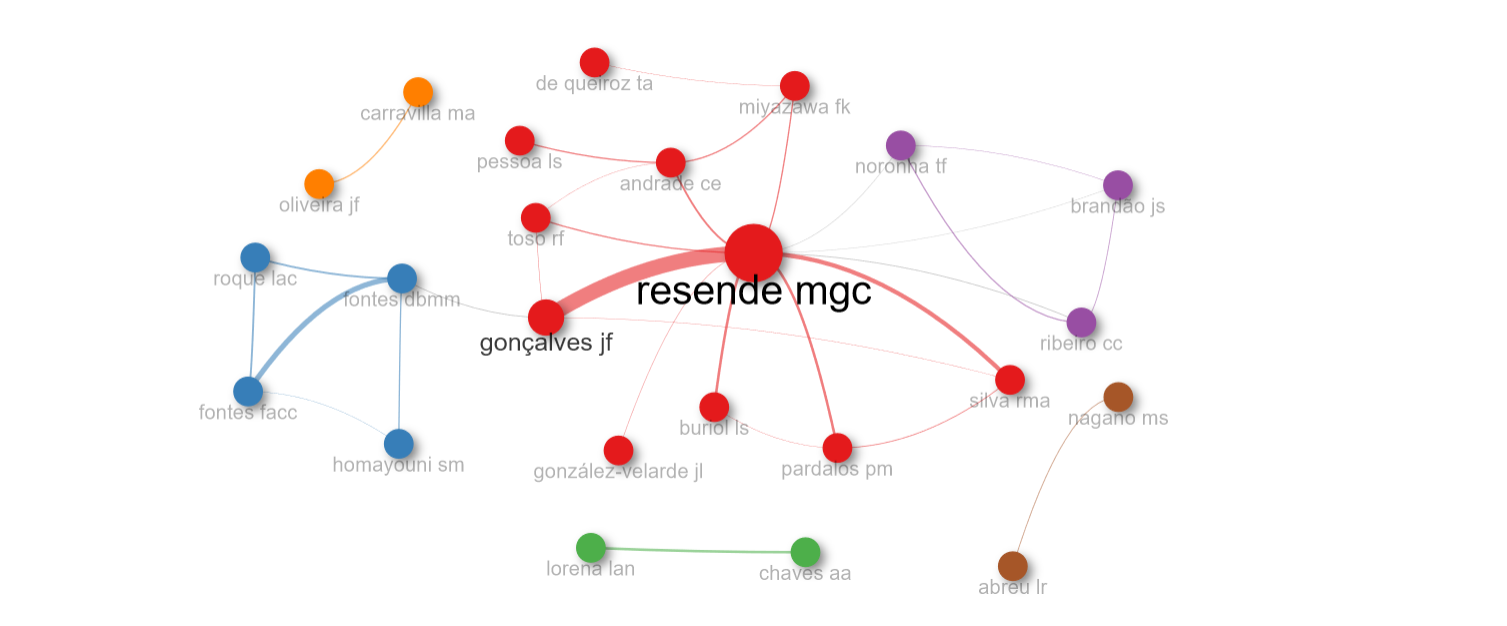}
    \caption{Collaboration network of the articles. The size of the nodes indicates a higher amount of articles, while the thickness of the edges indicates a higher amount of collaborations. Edges are only shown if the authors collaborated on more than two papers.}
    \label{Figure:Results:Co_author_network}
\end{figure}

One may also note that a significant amount of papers is co-authored by the first most prolific author and the others, especially the second most prolific (Gon\c{c}alves, J.F.). Moreover, Resende's most cited paper in the database is also Gon\c{c}alves' \citep{Goncalves2011:BRKGA} and is also the most cited paper in the database. There are only four single-authored papers, and international co-authorship happens in around 40\% of the works in the database.
The most significant co-authorship relations may be visualized in Figure~\ref{Figure:Results:Co_author_network}.

%
\begin{sidewaystable}[p]
    \centering
    \caption{10 authors with most published papers. Note that the total citations are calculated among the works in the database.}
        \begin{tabular*}{\textwidth}{@{\extracolsep\fill}lccll}
            \hline
            \textbf{Author} & \textbf{No. p.} & \textbf{Tot. Cit.} & \textbf{Affiliation(s)} & \textbf{Country} \\
            \hline
            Resende M. G. C. & 51 & 3223 & AT\&T Labs Research / University of Washington & USA\\
            Gon\c{c}alves J. F. & 30 & 2733 & Universidade do Porto & Portugal\\
            Fontes D. B. M. M. & 16 & 146 & Universidade do Porto & Portugal\\
            Andrade C. E. & 14 & 241 & Universidade de Campinas / AT\&T Labs Research & Brazil / USA\\
            Chaves A. A. & 13 & 190 & Universidade Federal de S\~ao Paulo & Brazil\\
            Ribeiro C. C. & 11 & 245 & Universidade Federal Fluminense & Brazil\\
            Pardalos P. M. & 10 & 379 & University of Florida & USA\\
            Silva R. M. A. & 10 & 105 & Universidade Federal de Pernambuco & Brazil\\
            Fontes F. A. C. C. & 10 & 64 & Universidade do Porto & Portugal\\
            Buriol L. S. & 8 & 232 & Universidade Federal do Rio Grande do Sul / Amazon.com & Brazil / USA\\
            \hline
        \end{tabular*}
    \label{Table:Results:Authors}
\end{sidewaystable}


Table~\ref{Table:Results:Affiliations} presents the more significant affiliations of the authors. One may note that the authors are mainly from the USA, Brazil, and Portugal. More than that, one must comment that the affiliations of authors with the most published titles are also the ones with more published works.

\begin{table}[htb]
    \centering
     \caption{15 institutes with most published papers.}
    \begin{tabular}{lll}
        \hline
        \textbf{Affiliation} & \textbf{No. papers} & \textbf{Country} \\
        \hline
        Universidade do Porto & 61 & Portugal\\
        AT\&T Labs Research & 38 & USA\\
        Universidade Federal Fluminense & 17 & Brazil\\
        Universidade Federal do Rio Grande do Sul & 16 & Brazil\\
        Universidade de S\~ao Paulo & 13 & Brazil\\
        Universidade Federal de São Paulo & 10 & Brazil\\
        Instituto Superior de Engenharia do Porto & 10 & Portugal\\
        Universidade Federal de Pernambuco & 10 & Brazil\\
        Universidade Federal do Rio de Janeiro & 10 & Brazil\\
        University of Florida & 10 & USA\\
        Universitat Polit\`ecnica de Catalunya & 9 & Spain\\
        Universidade de Campinas & 9 & Brazil\\
        Universidade Federal de Minas Gerais & 8 & Brazil\\
        University of Washington & 8 & USA\\
        Universidade Federal do Cear\'a & 6 & Brazil\\
        \hline
    \end{tabular}
    \label{Table:Results:Affiliations}
\end{table}


Table~\ref{Table:Results:Countries} presents the ten countries with the most published papers. One may note that this confirms the trend shown in Tables~\ref{Table:Results:Authors} and~\ref{Table:Results:Affiliations} regarding Brazil, the USA, and Portugal.

\begin{table}[htb]
    \centering
     \caption{10 countries with most published papers.}
    \begin{tabular}{lc}
        \hline
        \textbf{Country} & \textbf{No. papers}\\
        \hline
        Brazil & 215\\
        USA & 111\\
        Portugal & 94\\
        China & 29\\
        Spain & 28\\
        France & 21\\
        Italy & 15\\
        Germany & 14\\
        Mexico & 13\\
        Indonesia & 10\\
        \hline
    \end{tabular}
    \label{Table:Results:Countries}
\end{table}



\section{Co-citation analysis}
\label{Section:Results:Cocitation}

\begin{figure}
    \centering
    \includegraphics[scale = 0.5]{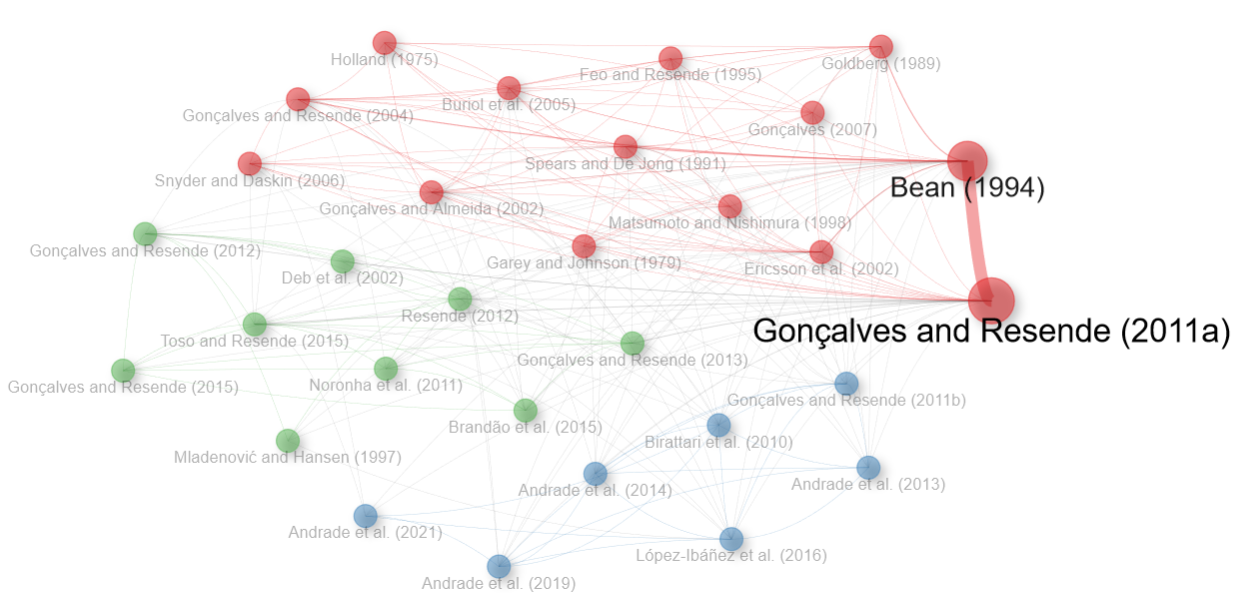}
    \caption{Co-citation network of the articles. The thickness of the edges indicates a stronger co-citation relationship.}
    \label{Figure:Results:co_citation_network}
\end{figure}

Co-citation analysis examines the relation between works cited together to present the most influential works in the researched area \citep{Persson2009:bibexcel}. Thus, it was used to answer the  RQ2. The co-citation network of the studied articles may be seen in Figure~\ref{Figure:Results:co_citation_network} for the~30 works with the most citations. The size of the circles indicates the number of citations, while the lines' thickness indicates the amount of co-occurrences.

The earliest study in this co-citation network is the book authored by \citet{Holland1975:genetic_algorithm}. In it, the author introduces the concept of genetic algorithms by presenting not only the theoretical foundations based on the Darwinian theory of evolution but also illustrating applications in different areas of knowledge.

Following it is the book of \citet{Garey1979:computers_intractability}. The authors study and define the theory of ``NP--completeness,'' emphasizing the concepts and techniques most useful for practical purposes. This book also presents an extensive list of at--the--time known NP--complete and NP--hard problems and their variants.

Following this, the book of \citet{Goldberg1989:genetic_algorithms} was published. This book explores the theory given in~\citet{Holland1975:genetic_algorithm}, being an introduction to all essential topics needed to use a genetic algorithm, including crossover, mutation, classifier systems, and fitness scaling. It also has a complete Pascal listing of a simple genetic algorithm.

The study of \citet{Spears1991:multi_point_crossover} makes a theoretical analysis of different multi-point crossovers, focusing on $n$-point and uniform crossovers. This work shows that the uniform crossover converges on average more slowly but to better solutions than other alternatives.

In the work of \citet{Bean1994:random_keys}, the concept of chromosomes composed of random keys was introduced. The author presents the advantages of such representation, such as the generalized mapping of solutions to search space, and explores several optimization problems with differing chromosome-to-solution mappings. The author also demonstrates that an algorithm composed of the operators of reproduction, uniform crossover with a probability of~0.7 between two randomly chosen chromosomes in the population, and random generation of new solutions in each iteration is very robust, with excellent results in certain classes of problems.

\citet{Feo1995:GRASP} introduce the greedy randomized adaptive search procedure (GRASP), a heuristic based on the interactive construction of solutions. This construction is made by randomly choosing the next move among the best candidates. Similarly, \citet{Mladenovic1997:vns} introduces the variable neighborhood search (VNS) algorithm, a local search method that sequentially explores several neighborhoods. In the literature, both GRASP and VNS approaches are frequently compared with BRKGA. 

The article of \citet{Matsumoto1998:mersenne_twister} presents the Mersenne Twister, a pseudo-random number generator used to obtain numbers between zero and one with uniform probability. This generator is as fast as the existing alternatives at the time, with a longer period and far larger distributions.

The work of \citet{Goncalves2002:hybrid_assembly_line_balancing} uses a hybrid genetic algorithm with random-key gene representation alongside a local search for the simple assembly line problem. The genetic algorithm uses an evolutionary strategy identical to the one used in~\citet{Bean1994:random_keys}, with a higher percentage of mutated chromosomes and a biased uniform crossover strategy similar to the one in~\citep{Spears1991:multi_point_crossover}.
In the study by \citet{Ericsson2002:Genetic_alg_OSPF}, a genetic algorithm for the weight setting problem in OSPF routing is detailed. This genetic algorithm does not use random keys in gene representation but in the crossover procedure to select which genes shall be inherited. In this case, a biased coin toss is performed, favoring the elite parent. Together, both works are some of the first methods with BRKGA characteristics.

\citet{Deb2002:NSGAII} introduces the Non-dominated Sorting Genetic Algorithm II (NSGA-II), a multi-objective evolutionary algorithm with a fast non-dominated sorting approach and diversity-maintaining strategies. This method is one of the most effective for multi-objective problems and is frequently the benchmark in this category.

\citet{Goncalves2004:manufacturing_cell_formation} use a hybrid genetic algorithm with local search to solve the manufacturing cell formation problem. This algorithm uses random-key gene representation, with an extra gene to indicate the number of cells to be formed, and an evolutionary strategy identical to the one used in~\citet{Goncalves2002:hybrid_assembly_line_balancing}.

\citet{Buriol2005:weight_setting_problem_OSPF_routing} extends the work of~\citet{Ericsson2002:Genetic_alg_OSPF}, with similar chromosome representation and crossover scheme, and a local improvement procedure incorporated after the crossover step.

\citet{Snyder2006:generalized_tsp} study a generalized traveling salesman problem with a random-key genetic algorithm. This approach uses a novel combination of random-keys representation, genetic algorithms, and improvement methods such as local search heuristics. They also use an elitist reproduction operator and a parameterized uniform crossover \citep{Spears1991:multi_point_crossover} but with randomly chosen parents among the whole population. The method is competitive compared to others from the literature, is simple to implement, and can be easily adapted to other problem characteristics. 

The work of \citet{Goncalves2007:two_dimensional_orthogonal_packing} hybridizes a placement procedure with a random-key genetic algorithm for a two-dimensional orthogonal packing problem. The author uses an evolutionary strategy similar to the one in~\citet{Goncalves2002:hybrid_assembly_line_balancing}.

In the article of \citet{Birattari2010:f_race_iterated}, the authors propose a new method for selecting parameter configurations for algorithms. This method, called iterated F-race, is based on discarding non-promising parameter configurations as soon as there is statistical proof that they are sub-optimal. This way, the algorithm focuses on the best configurations instead of spending time and computational effort on uninteresting ones. As it examines all parameters together, it is also an interesting option when one has a high number of parameters to select.

\citet{Noronha2011:routing_wavelenght_assignment} uses a random-key-based genetic algorithm to solve the routing and wavelength assignment problem. This heuristic approach extends one of the best strategies present in the literature for this problem by adapting it to a multi-start and evolutionary framework. The evolutionary strategy is similar to the one used in~\citet{Goncalves2002:hybrid_assembly_line_balancing}.

The article of \citet{Goncalves2011:BRKGA} formally presents the biased random-key genetic algorithm as a class of heuristics that considers random-key-based chromosome representation alongside elitist and biased uniform crossover. It also presents possible decoders for several optimization problems and comparisons with other heuristics to show the competitiveness of BRKGA.

In the study of \citet{Goncalves2011:multi_pop_constrained_2d_orthogonal_packing}, the authors use a multi-population random-key based genetic algorithm to solve a two-dimension orthogonal packing problem. This algorithm uses a similar evolutionary strategy as in \citet{Goncalves2004:manufacturing_cell_formation}. In this case, more than one population runs independently, and those populations exchange two of their best chromosomes after a set amount of generations. Also, alongside randomly generated chromosomes, the populations are initiated with four known solutions, something proved by the results to increase the quality of the obtained solutions.

\citet{Resende2012:BRKGA_telecom} details several applications of BRKGA in five telecommunication problems: weight-setting problem in OSPF routing, design of OP networks when routing is done with OSPF, location of redundant servers, location of signal regenerators in optical networks, and routing and wavelength assignment in optical networks. For each problem, the decoding strategies are presented, and the results are compared favorably with other algorithms.

The article of \citet{Goncalves2012:multi_pop_container_loading} presents a parallel multi--population BRKGA for the single container loading problem. The chromosome considers the packing sequence and the layer types, i.e., in which layers to pack each package. Two variants of the BRKGA are compared with others from the literature and show that the algorithm performs very well in all types of instances and has the best overall results from all approaches in the literature.

In the work of \citet{Goncalves2013:2d_3d_bin_packing}, a BRKGA for 2D and 3D bin packing problems is presented. The BRKGA was implemented with parallelization of the decoding process and a chromosome representation that considers each item's packing order and orientation. The results indicate that the BRKGA consistently outperforms other existing algorithms.

\citet{Andrade2013:Evolutionary_Algorithm_kIMDMTSP} detail a BRKGA with local search inside the decoder for the $k$-interconnected multi-depot multi-traveling salesman problem. The authors compare the BRKGA with a simple multi-start heuristic (MSH) with the same local search for different categories of instances. Overall, the BRKGA appears to perform significantly better than the MSH.

In the article by \citet{Andrade2014:Evol_Alg_Overlapping_Correlation_Clustering}, the authors explore four variants of BRKGA for overlapping correlation clustering. The four variants consider two possible chromosome representations, namely compact and extended, and two local searches, one for error reduction and the other based on the literature. The results show that the extended representation with literature-based local search performs better than its competitors but also indicate high convergence times for all BRKGA variants.

\citet{Toso2015:C++_app_BRKGA} presents BRKGA API, in which all problem-independent components of the BRKGA framework are implemented. It also gives users all the tools needed to develop a decoder procedure correctly, which may be used in single- and parallel-threads.

In the work by \citet{Goncalves2015:unequal_area_facility_layout}, an unequal area facility layout problem, a version of the quadratic assignment problem, is studied. The authors hybridized a BRKGA with linear programming constraints, being one of the first hybridizations of this method and exact methodologies. The hybrid algorithm outperformed several others from the literature in the experiments. 

The article of \citet{Brandao2015:single_round_load_scheduling} studies the single-round divisible load scheduling problem and proposes a BRKGA that outperformed some strategies for the literature.

The work of \citet{LopezIbanez2016:irace} introduces the ``irace'' package, in which the iterated F-race of~\citep{Birattari2010:f_race_iterated} is implemented. The authors detail several of the package options and offer examples of its usage in diverse areas of knowledge.

\citet{Andrade2019:flowshop_scheduling} study a flow shop scheduling problem and propose a new feature for BRKGA called the ``shaking'' operator. This is a partial restart of the population by modifying the elite solutions and randomly re-initializing the non-elite set. This way, some of the benefits of evolution are retained while diversity is increased. The authors show the effectiveness of this operator in solving the problem.

Finally, \citet{Andrade2021:BRKGA_MP_IPR} introduces the BRKGA with multi--parent crossover and path-relinking intensification strategy, called BRKGA-MP-IPR. The authors present an API for the method and detail its several parameters, functions, and variants.

One can note some similarities among the works present in the co-citation analysis. First, we have the studies that introduce the theoretical background of BRKGA, with genetic algorithms \citep{Holland1975:genetic_algorithm,Goldberg1989:genetic_algorithms}, random-keys \citep{Bean1994:random_keys}, parallelized crossover \citep{Spears1991:multi_point_crossover}, random number generator \citep{Matsumoto1998:mersenne_twister}, and the theory of NP-completeness \citep{Garey1979:computers_intractability}. Second, the methods that are often used in experiments as benchmarks, such as GRASP \citep{Feo1995:GRASP}, VNS \citep{Mladenovic1997:vns}, and NSGA-II \citep{Deb2002:NSGAII}. Third, the works with initial characteristics of BRKGA \citep{Goncalves2002:hybrid_assembly_line_balancing,Ericsson2002:Genetic_alg_OSPF,Goncalves2004:manufacturing_cell_formation,Buriol2005:weight_setting_problem_OSPF_routing,Snyder2006:generalized_tsp} and their variants \citep{Goncalves2007:two_dimensional_orthogonal_packing,Noronha2011:routing_wavelenght_assignment,Goncalves2011:multi_pop_constrained_2d_orthogonal_packing}. Fourth, the novel applications and improvements of the framework \citep{Goncalves2011:BRKGA,Resende2012:BRKGA_telecom,Goncalves2012:multi_pop_container_loading,Goncalves2013:2d_3d_bin_packing,Andrade2014:Evol_Alg_Overlapping_Correlation_Clustering,Toso2015:C++_app_BRKGA,Goncalves2015:unequal_area_facility_layout,Brandao2015:single_round_load_scheduling,Andrade2019:flowshop_scheduling,Andrade2021:BRKGA_MP_IPR}. Fifth and last, the works with methods for parameter selection \citep{Birattari2010:f_race_iterated,LopezIbanez2016:irace}.


\section{Co-word analysis}
\label{Section:Results:Coword}

\begin{figure}[htb]
    \centering
    \includegraphics[scale = 0.3]{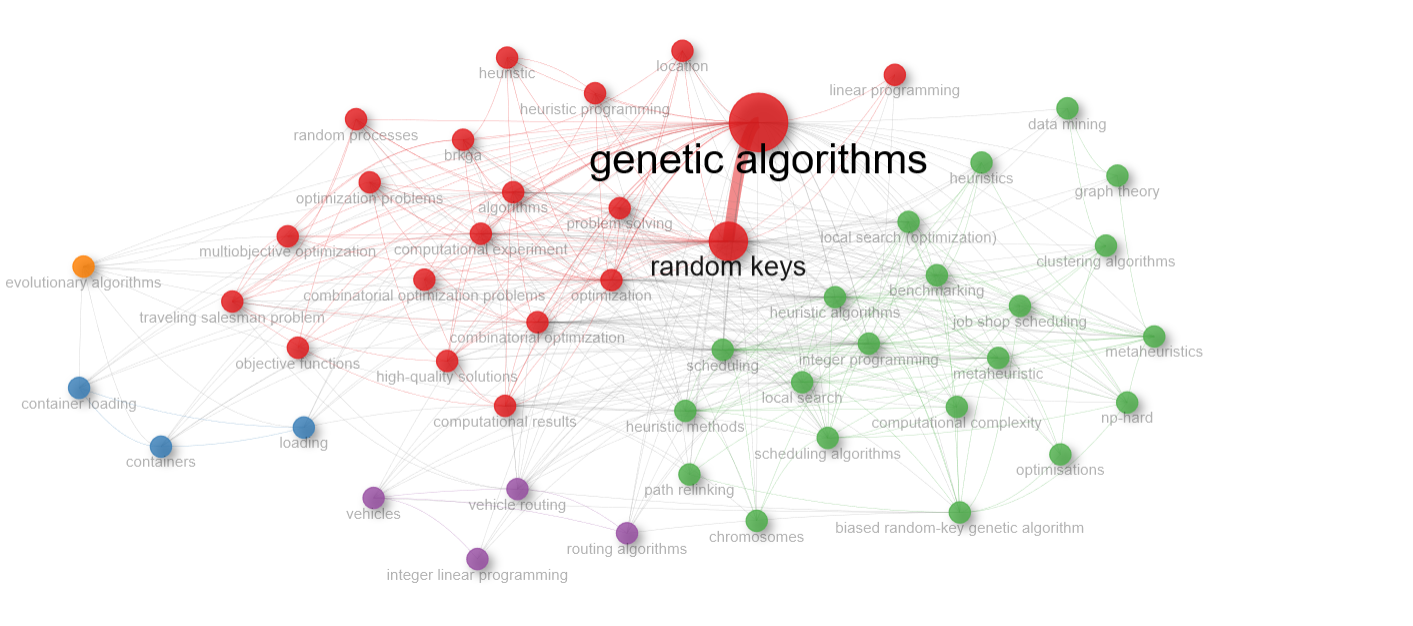}
    \caption{Co-occurrence network for the keywords observed in the works.}
    \label{Figure:Results:Co_occurence_net}
\end{figure}

Co-word analysis intends to answer RQ3, ``Which are the main themes present in BRKGA studies and how did they evolve?''. In Figure~\ref{Figure:Results:Co_occurence_net}, one may observe the co-occurrence network of the keywords observed in all articles. In it, we have a cluster centered on both genetic algorithms and random keys, another in combinatorial optimization and scheduling, a third in vehicle routing, and a fourth in container loading. Those are also some of the most frequent keywords seen in the database.

\begin{figure}[htb]
    \centering
    \includegraphics[scale = 0.35]{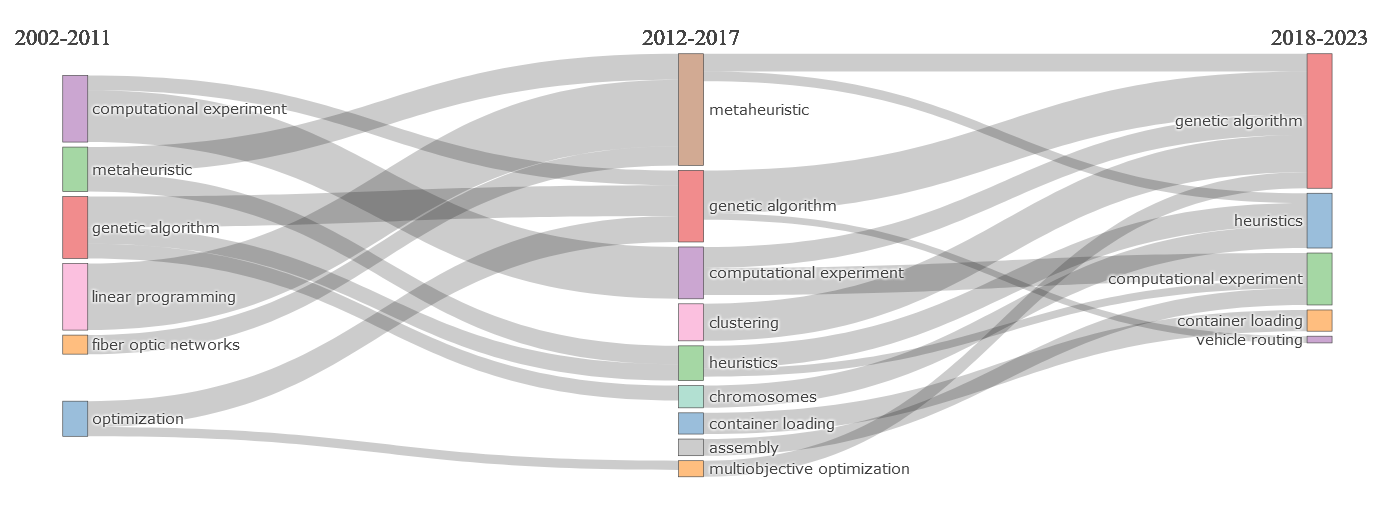}
    \caption{Thematic evolution map of the three periods.}
    \label{Figure:Results:ThematicEvolution}
\end{figure}

To better understand and have a more in-depth analysis, we divided the documents into three successive periods: 2002-2011, 2012-2017, and 2018-2023, with~28,~73, and~152 papers, respectively.
The thematic evolution map of the periods can be seen in Figure~\ref{Figure:Results:ThematicEvolution}. The association strength between words can be seen by the thickness of the lines, calculated by the inclusion index. Meanwhile, the size of the cluster rectangle indicates the amount of occurrences of the keywords associated with that cluster. For example, the \textit{genetic algorithm} cluster in the first period has~111 frequency, as measured by 1000 documents, while \textit{fiber optic networks} only has~28.6.

Note that, in the first period, the cluster of keywords labeled \textit{linear programming} is the most frequently seen, followed by \textit{genetic algorithm} and \textit{computational experiment}. For the second period, \textit{metaheuristic} becomes the most frequent cluster of keywords, closely followed by \textit{genetic algorithm} and \textit{computational experiment}. Finally, in the most recent period, \textit{genetic algorithms} becomes the biggest cluster, followed by \textit{heuristics} and \textit{computational experiment}. The keywords pertaining to themes \textit{genetic algorithm} are among the most frequent in all clusters, with it being the most frequent theme overall. As all papers are expected to deal with the BRKGA algorithm, which is a genetic algorithm, the frequency of this theme in the periods is expected. Another often-seen theme is \textit{computational experiment}, something also expected due to the restriction of only papers that use BRKGA to solve an optimization problem.

\begin{figure}[htb]
    \centering
    \includegraphics[scale = 0.3]{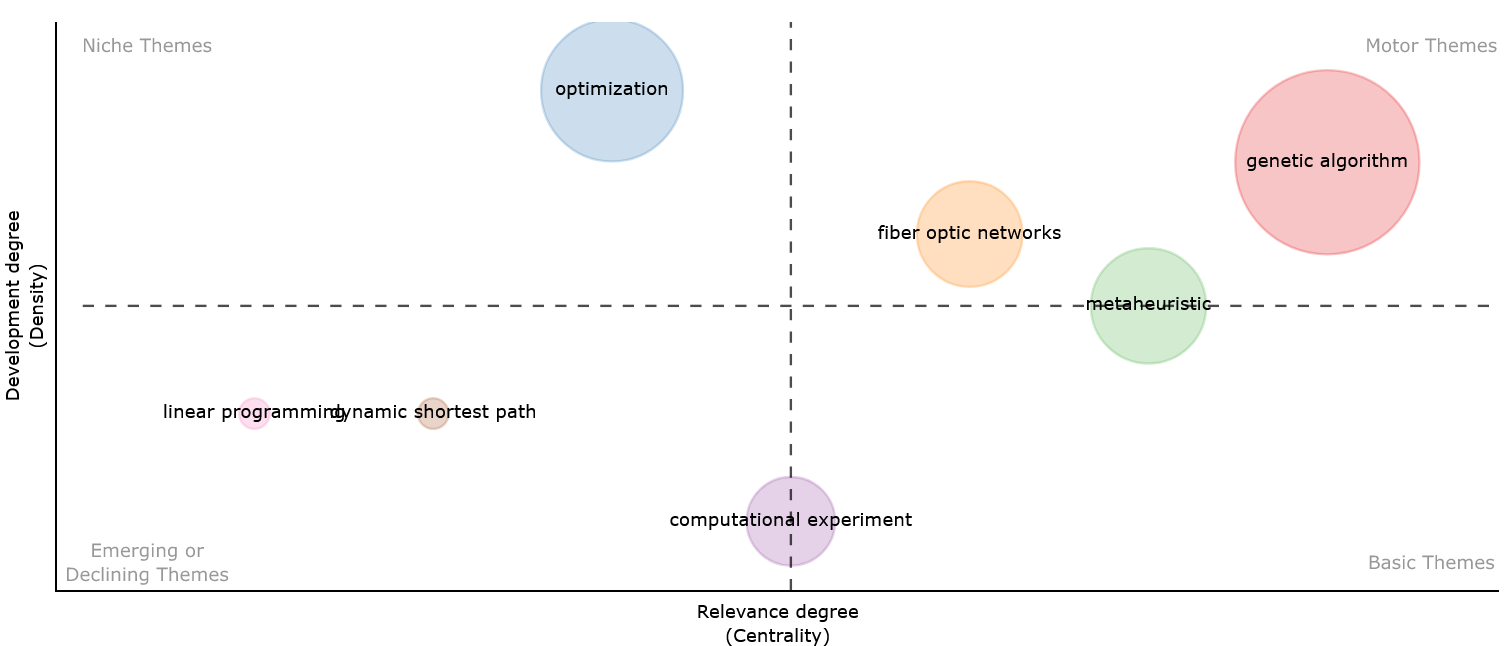}
    \caption{Strategic Diagram for 2002-2011 articles.}
    \label{Figure:Results:Strategic_pre2011}
\end{figure}

\begin{figure}[htb]
    \centering
    \includegraphics[scale = 0.25]{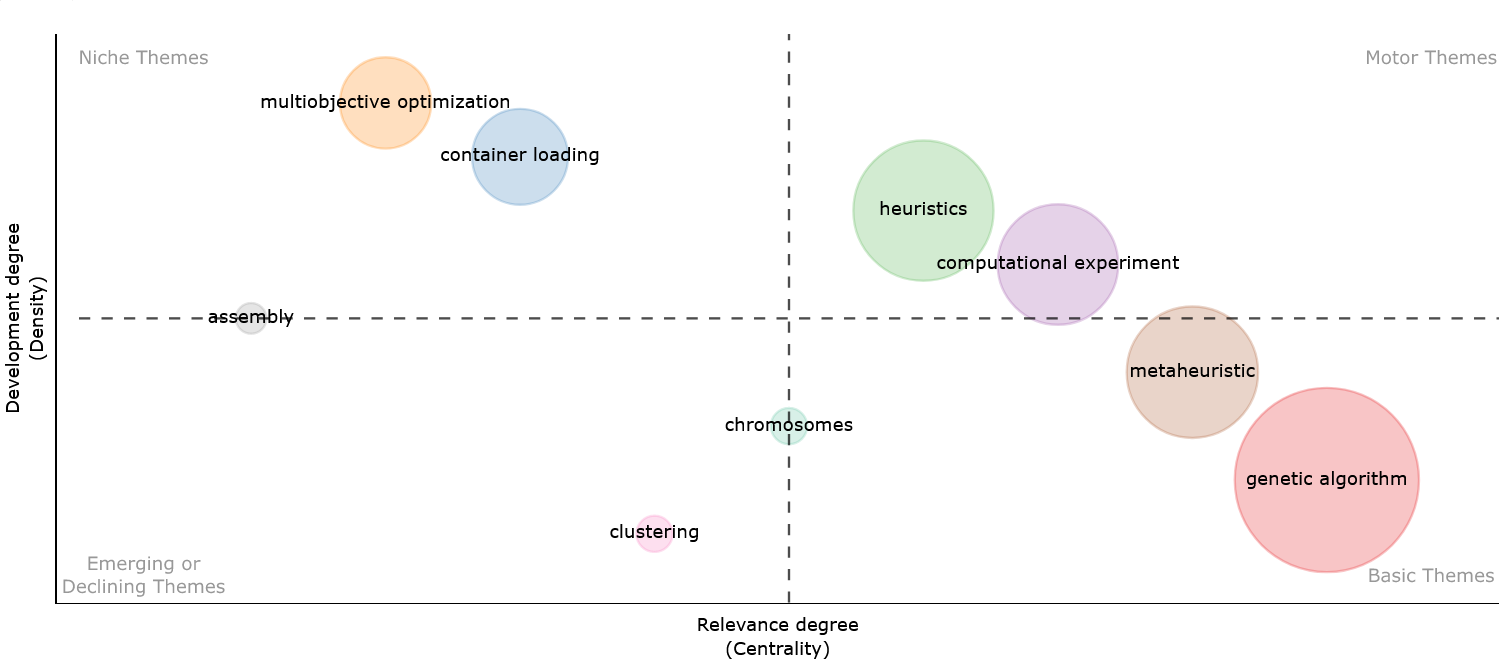}
    \caption{Strategic Diagram for 2012-2017 articles.}
    \label{Figure:Results:Strategic_2011_2017}
\end{figure}

\begin{figure}[htb]
    \centering
    \includegraphics[scale = 0.25]{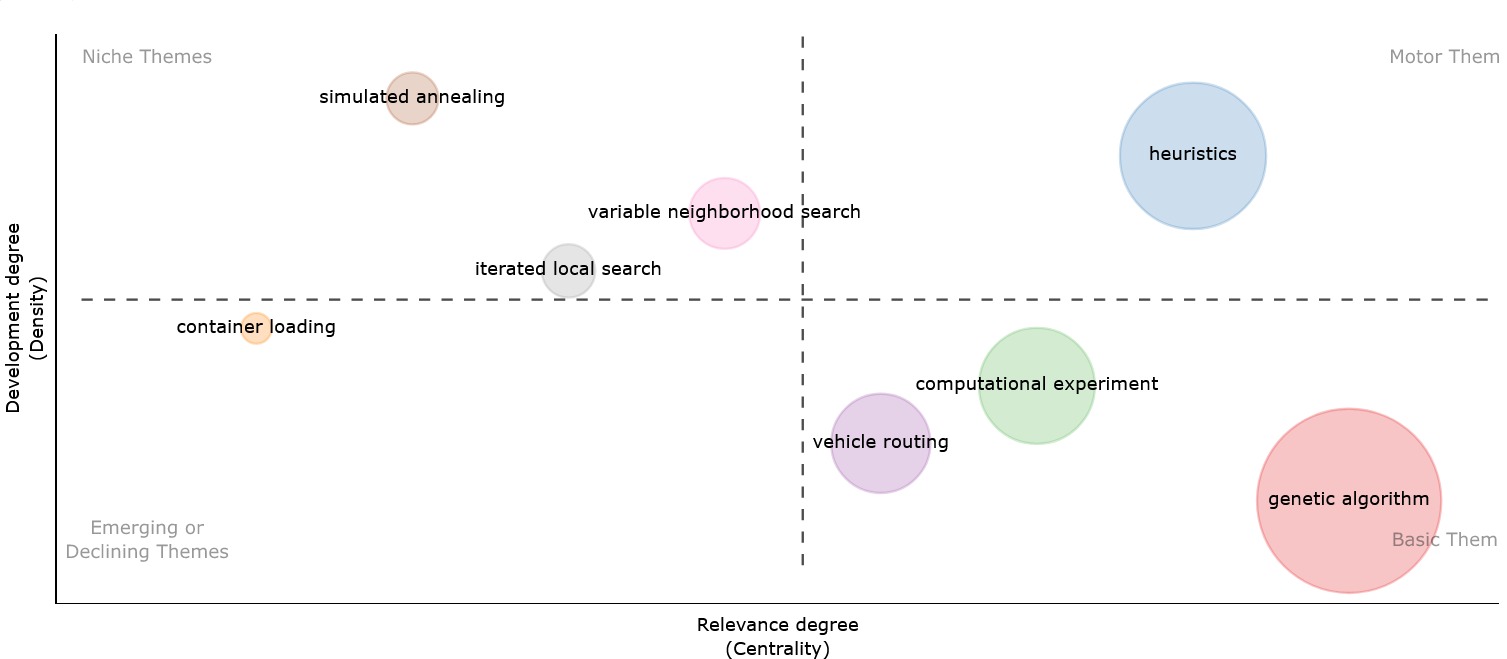}
    \caption{Strategic Diagram for 2018-2023 articles.}
    \label{Figure:Results:Strategic_2018_2023}
\end{figure}

Figures~\ref{Figure:Results:Strategic_pre2011}, \ref{Figure:Results:Strategic_2011_2017} and~\ref{Figure:Results:Strategic_2018_2023} present the strategic diagram for the three periods. In them, the thickness of the circles indicates the number of papers with the keywords that belong to that cluster. Note that the labels indicate the most frequent keyword of that cluster.

In the 2002-2011 period, \textit{genetic algorithm} and \textit{fiber optic networks} were core or motor themes, meaning that they were highly related and quoted themes in the database. \textit{Genetic algorithm} also has the highest concentration of papers among the themes of this period. The theme \textit{metaheuristic} was also highly relevant, but not as well developed.

Among the most significant papers with \textit{genetic algorithms} are \citet{Goncalves2002:hybrid_assembly_line_balancing} and \citet{Goncalves2011:BRKGA}. The former introduces the framework that later would become BRKGA, with biased uniform crossover, mutation operator, and random-key chromosome representation, while the latter formally presents the BRKGA framework. Meanwhile, \textit{fiber optic networks} and related keywords may be seen in \citet{Ruiz2011:WSON_multilayer_network}, which studies the survivable IP/MPLS--over--WSON multilayer network optimization problem. \citet{Festa2010:grasp_tuning_brkga} uses a BRKGA to tune the parameters of a GRASP with path-relinking, being one of the papers in theme \textit{metaheuristic}.

\textit{Optimization}, meanwhile, is a niche or isolated theme. That means it has a high density or development but is not strongly correlated with the other themes. One study with this theme is \citet{Goncalves2011:lot_sizing_scheduling_backorders}, where a BRKGA is customized for the economic lot scheduling problem. This approach is hybridized with an LP model and had good performance in experiments with randomly generated instances. 

The cluster labeled with \textit{computational experiment} denotes a highly promising theme of this period, as it has a high correlation with other themes with low development. One such work with this theme is \citet{Mendes2009:resource_constrained_scheduling}. In it, the authors present the results of a BRKGA used in a resource-constrained scheduling problem.

\textit{Dynamic Shortest Path} and \textit{linear programming} are the emerging or declining themes of this period. In \citet{Buriol2010:road_congestion}, the dynamic shortest path is used to update a solution when there is a modification on the road network. Similarly, in \citet{Reis2011:OSPF_DEFT_routing_network_congestion}, the dynamic shortest path was adapted from the one in \citet{Buriol2005:weight_setting_problem_OSPF_routing}, in which it is used to recompute shortest paths after modifications of link weights. A linear programming model is compared with the results of a proto-BRKGA in \citet{Ericsson2002:Genetic_alg_OSPF}.

For the second period, \textit{computational experiment} and \textit{heuristics} are motor themes. It is expected, as all papers in the database used a metaheuristic method, BRKGA, in computational experiments. Among the works that use these themes, we have \citet{Duarte2014:regenerator_location}, which uses hybrid heuristics and metaheuristics to solve the regenerator location problem. A hybrid GRASP with path-relinking is proven to be the best approach among the ones used by the authors.  For \textit{computational experiment}, we have \citet{Heilig2016:resource_management_cloud}. This paper studies a cloud resource management problem using a hybrid BRKGA with a local search.

There are three niche themes in this period: \textit{multiobjective optimization}, \textit{container loading}, and \textit{assembly}. One work with a \textit{multiobjective optimization} problem is \citet{Tangpattanakul2015:scheduling_observations_satellite}, which proposes a bi-objective BRKGA approach with modifications on the chromosome ordering phase. A 3D \textit{container loading} problem is studied by \citet{Goncalves2012:multi_pop_container_loading}, with a multi-population BRKGA that had better performance than several other approaches. The study from \citet{Moreira2012:assembly_line_worker_assignment_balancing} introduces an \textit{assembly} line worker assignment and balancing problem. Their hybrid BRKGA with local search showed better performance than several other approaches from the literature.

\textit{Clustering} is the emerging theme of this period. It may be seen in \citet{Festa2013:data_clustering}, which uses a BRKGA for clustering biological data. 

Finally, the promising themes for this second period are \textit{genetic algorithm}, \textit{metaheuristic}, and \textit{chromosome}.  Among those that count the words among their keywords, we have the aforementioned \citet{Goncalves2012:multi_pop_container_loading}, where variants of the BRKGA are compared favorably with others from literature with regards to solution quality. The paper of \citet{Silva2015:library_BRKGA} is also present on these themes and introduces a Python/C++ library for BRKGA.  In \citet{Andrade2015:winner_auctions}, the authors introduce six versions of BRKGA for the winner determination problem in combinatorial auctions. Three of the BRKGA variants use solutions of intermediate linear programming relaxations of an exact mixed integer-linear programming model as initial chromosomes of the populations.

For the 2018-2023 period, the theme of \textit{heuristics} became the core theme, with an increase in both development and relevance. \citet{Mauri2021:multiproduct_facility_location} studies a two-stage multi-product capacitated facility location problem with hybrid BRKGA+LS that outperforms search-based \textit{heuristic} algorithms.

\textit{Simulated annealing}, \textit{variable neighborhood search}, and \textit{iterated local search} are niche themes of this period, with low centrality but high development. These are algorithms that are frequently compared or hybridized with BRKGA in the experiments. \citet{Andrade2019:scheduling_software_cars} compares their approach with several, including \textit{simulated annealing} an \textit{iterated local search}. ILS variants were also noted in \citet{Huang2018:road_network_disruptions,Rocholl2021:scheduling_parallel_common_due_date}. \citet{Silva2019:multicommodity_tsp} uses a hybrid BRKGA with ILS, which is then hybridized with VNS. \citet{Monch2018:matheuristic_batch_scheduling} shows that BRKGA is a better method than VNS for a parallel batch processing machine problem. 

As an emerging theme, we have \textit{container loading}. A new methodology for the container loading problem is studied in \citet{Ramos2018:container_loading}. The authors treat load balance as an important constraint for the problem, something normally simplified in literature. The load balance is considered a vehicle-specific constraint in this new formulation. The new approach was tested on 1,500 instances and showed an improvement in solution quality and algorithm performance.

At last, \textit{vehicle routing}, \textit{computational experiment}, and \textit{genetic algorithm} are the basic themes for this period. This means a decrease in development for the second theme and a slight increase in relevance for the third. One can observe the first theme in \citet{Schenekemberg2022:dial_a_ride}, which explores the dial-a-ride problem with private fleet and common carrier. The proposed hybrid BRKGA with local search successfully outperformed other methods in a case study. We can cite \citet{Andrade2019:flowshop_scheduling} for \textit{computational experiment}, in which several heuristics for a flowshop scheduling problem are presented. The authors detail the results for several methods on experiments made with 120 test instances, including an algorithm based on iterated local search, iterated greedy search, and BRKGA. This paper also introduces the shaking mechanism, in which the population is partially restarted, and proves its effectiveness inside the BRKGA framework.  Among the studies that cite \textit{genetic algorithm}, \citet{Fadel2021:statistical_disclosure_control} use genetic algorithms such as BRKGA for statistical disclosure control. The context of this method is the sharing of data between public and private organizations, that must be done without compromising confidentiality or losing important data characteristics. The algorithms, thus, must both cluster and aggregate the data so that the statistical measures, i.e. mean or standard deviation, are not modified. The experiments point out that the genetic algorithms are adequate to produce good results, however with high run times.

\section{Conclusion}
\label{Section:Conclusion}

This paper conducted a reproducible systematic literature review encompassing~253 articles of applications of the Biased Random-Key Genetic Algorithm (BRKGA). Although the initial applications of this methodology emerged in the early 2000s \citep{Goncalves2002:hybrid_assembly_line_balancing,Ericsson2002:Genetic_alg_OSPF} , the formal framework was not established until~2011 \citep{Goncalves2011:BRKGA}. Since its inception, this metaheuristic has found extensive application across various optimization problems, demonstrating its reliability, efficiency, and versatility.

This paper was guided by three research questions (RQ). Citation analysis was used to answer the first RQ: \textit{Who are the most influential researchers for this algorithm?} The citation analysis shows that the most influential authors are Resende, M.G.C., and Gon\c{c}alves, J.F., both relative to the number of papers and total of citations inside the database. The most cited paper among the~253 selected is co-authored by both \citep{Goncalves2011:BRKGA}, and there is a strong correlation between the number of papers and the total of citations in the database. The analysis also points out that Brazil, the USA, and Portugal are, by far, the countries with the most authors publishing BRKGA. This can be explained by the relationship between the two most influential authors and researchers based in these countries.

The second research question, \textit{What are the most influential papers for the BRKGA framework?}, was answered with a co-citation analysis. The papers are presented in chronological order and point out that papers focused on the theoretical parts of the BRKGA framework, such as multi-point crossover, random-keys, and elitist strategy, are some of the most frequently co-cited works. There is also a strong presence of application articles and methods for parameter selection. This analysis presents the backbone of seminal works on BRKGA and is a chronological guide for any researcher who desires to study this metaheuristic.

The last RQ, \textit{Which are the areas of application of BRKGA and how did they evolve?}, was answered with a co-word analysis, which presents the contemporary research themes and their evolution, and is divided into three time periods. In the first period (2002-2011), the main themes were genetic algorithms and fiber optic networks. These point to the majority of the applications in the time--period, and the fact that the metaheuristic was only a variation of GA at the time, without a specific name. In the second period (2012-2017), heuristics and computational experiments were the motor themes of the studies, while heuristics remained the most relevant and developed area in the most recent period (2018-2023).

Potential approaches to expand the scope of this review include incorporating additional databases like Web of Science and arXiv, as well as encompassing theses, dissertations, and non-peer-reviewed materials such as technical reports and preprints.


\section*{Acknowledgements}

This work was supported by the 
the Brazilian National Council for Scientific and Technological Development (CNPq) under Grant [number 312212/2021-6]; Brazilian Coordination for the Improvement of Higher Level Personnel (CAPES) under Grants [number 001 and 88887.815411/2023]; the Carlos Chagas Filho Research Support Foundation of the State of Rio de Janeiro (FAPERJ) under Grants [numbers 211.086/2019, 211.389/2019, and 211.588/2021].


\bibliography{bibfiles/revBRKGA,bibfiles/bibliometric,bibfiles/new_papers}



\end{document}